\documentclass[letterpaper, 10 pt, conference]{ieeeconf}  

\IEEEoverridecommandlockouts                              

\usepackage{graphicx}
\usepackage{graphics} 
\usepackage{mathptmx} 
\usepackage{times} 
\usepackage{amsmath} 
\usepackage{amssymb}  
\usepackage{xcolor}
\usepackage{subcaption} 

\usepackage{tikz}
\usetikzlibrary{shapes,arrows}
\usepackage{amsmath,bm,times}
\usepackage{amssymb}
\usepackage{verbatim}
\usetikzlibrary{calc,positioning,shapes.geometric}
\usetikzlibrary{backgrounds,fit,bayesnet,arrows}

\usepackage{pgfplotstable}
\usepackage{booktabs} 
\usepackage{rotating}
\usepackage{multirow} 
\usepackage{array} 

\usepackage{algorithm}
\usepackage{algorithmic}

\newcommand{\learnedLoss}{\mathcal{L}_{\text{learn}_\phi}}
\newcommand{\taskLoss}{\mathcal{L}_\text{MSE}}

\newlength\myindent
\makeatletter
\newcommand\fs@spaceruled{\def\@fs@cfont{\bfseries}\let\@fs@capt\floatc@ruled
  \def\@fs@pre{\vspace{0.5\baselineskip}\hrule height.8pt depth0pt \kern2pt}%
  \def\@fs@post{\kern2pt\hrule\relax}%
  \def\@fs@mid{\kern2pt\hrule\kern2pt}%
  \let\@fs@iftopcapt\iftrue}
\makeatother

\title{\LARGE \bf
Learning State-Dependent Losses for Inverse Dynamics Learning
}

\author{Kristen Morse$^{*}$, Neha Das$^{*}$, Yixin Lin, Austin S. Wang, Akshara Rai, Franziska Meier%
 \thanks{*authors contributed equally}%
 \thanks{All authors are with Facebook AI Research. \{kristenmorse, dasneha, yixinlin, wangaustin, akshararai, fmeier\}@fb.com}%
}

\pgfplotsset{compat=1.14} 
\begin{document}

\maketitle
\thispagestyle{empty}
\pagestyle{empty}

\begin{abstract}

Being able to quickly  adapt to changes in dynamics is paramount in model-based control for object manipulation tasks. In order to influence fast adaptation of the inverse dynamics model's parameters, data efficiency is crucial. Given observed data, a key element to how an optimizer updates model parameters is the loss function. In this work, we propose to apply meta-learning to learn structured, state-dependent loss functions during a meta-training phase. We then replace standard losses with our learned losses during online adaptation tasks. We evaluate our proposed approach on inverse dynamics learning tasks, both in simulation and on real hardware data. In both settings, the structured and state-dependent learned losses improve online adaptation speed, when compared to standard, state-independent loss functions.

\end{abstract}


\section{INTRODUCTION}
\begin{figure}
    \centering
    \includegraphics[
    width=0.52\columnwidth]{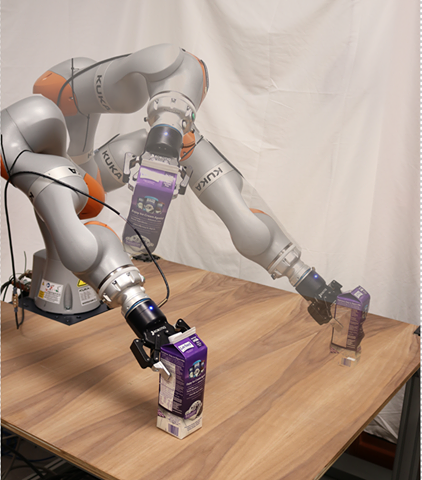}
    \caption{Picking and placing a milk carton with KUKA.}
    \label{fig:teaser-image}
    \vspace{-0.5cm}
\end{figure}

Truly autonomous robots need to be able to adapt their internal models when interacting with the environment. For instance, a robot that has learned its own dynamics needs to be able to adapt to the changes in dynamics caused by picking and placing a heavy object (Figure \ref{fig:teaser-image}). Humans are remarkably good at adapting to such changes very quickly. And studies on how humans adapt to external forces suggest that we \emph{learn how to adapt} to such changes in dynamics, and that we remember the adaptation strategy when encountering previously experienced perturbations \cite{Herzfeld:2014es}.

The larger goal of our work is to learn representations that endow robots with similar capabilities. In this work, we specifically look at fast learning or adaptation of inverse dynamics for model-based control. While many robots have analytical inverse dynamics models, they are often crude approximations to the actual dynamics of the robot. Furthermore, even if the models would be accurate enough for the robot's dynamics, they do not model the changes in dynamics caused by interaction with the environment. Thus, a lot of research has gone into data-driven methods for model-based control \cite{nguyen2011model}, with the hope that learning-based methods can either learn better models altogether \cite{nguyen2009local, schaal1998constructive}, or learn error-models on top of existing analytical ones \cite{ratliff2016doomed}.

In that context, online learning or adaptation of inverse dynamics models has been studied \cite{sigaud2011line,meier2016towards, gijsberts2013real}. When such online learning methods are deployed on real robots, one has to be concerned with computational and data efficiency of the optimizer. An online learner needs to make the most out of the recently observed data with as few parameter updates as possible. A key insight of our work is that standard losses, such as the Mean-Squared-Error (MSE), are not state-dependent, but observed errors and how much we should correct for them can be state-dependent. For instance, an optimizer could update dynamics parameters aggressively throughout most of the state-space, but needs to be more conservative around singular configurations, or after overcoming the effects of static friction. 

Learned, state-dependent losses could potentially incorporate such issues and thus lead to faster inverse dynamics model learning. Thus, in this work, we aim at improving data efficiency, by replacing standard loss functions for inverse dynamics learning, with learned, state-dependent losses.

Towards this goal, we develop a meta-learning framework, that trains loss functions for fast adaptation of inverse dynamics models. Our approach builds on \cite{bechtle2019meta}, a gradient-based meta-learning framework for learning loss functions. This work proposed learning losses that are parameterized as neural networks for various learning tasks. Here, we contribute the following: 1) we utilize \cite{bechtle2019meta} to formulate an offline meta-training algorithm for training loss functions for inverse dynamics learning, and 2) we propose two structured loss representations specifically suitable for learning inverse dynamics models, one of which is state-dependent.

We evaluate our proposed loss representations on several inverse dynamics learning tasks, both in simulation as well as on real hardware data of a 7 degree of freedom robot arm. Our results show that meta-learning structured losses leads to more robust meta-training of the losses, as compared to using typical neural network representations for the loss function. Furthermore, we find that our structured losses lead to faster learning and online adaptation of inverse dynamics models. Finally, especially on real robot learning experiments, the proposed state-dependent loss outperforms all other loss representations in terms of adaptation speed, providing experimental validation for the thought that learning state-dependent loss functions can improve data efficiency in inverse dynamics learning.

\section{Background and Related Work}
We start out by describing the control architecture used in this paper, and then summarize related works with respect to learning inverse dynamics and meta-learning.
\subsection{Background: Inverse Dynamics Control}
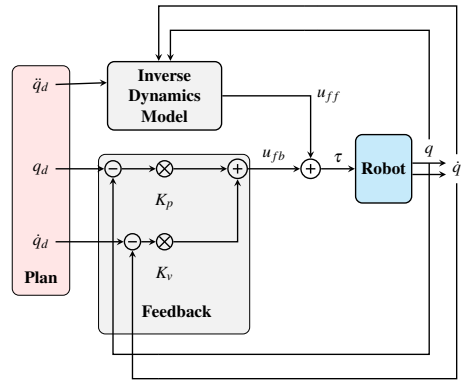
\begin{figure}[t]
    \centering
    \vspace{1mm}
    \tikzstyle{block} = [text centered, rounded corners, minimum height=4em, draw]
    \tikzstyle{smallblock} = [rectangle, draw, 
    text width=3em, text centered, rounded corners, minimum height=3em]
    \tikzstyle{robot}=[block, fill=cyan!20]
    \tikzstyle{plan}=[block, fill=red!10]
    \tikzstyle{inverse_dynamics}=[block, fill=gray!10]
    \tikzstyle{feed_back}=[block, fill=gray!10, minimum width=5em]
    \tikzstyle{feed_back_comp}=[smallblock, fill=gray!30, minimum size=2em]
    \tikzstyle{line} = [draw, -latex]
    \tikzset{PLUS/.style={draw,circle,append after command={
        [shorten >=\pgflinewidth, shorten <=\pgflinewidth,]
        ([yshift=-3mm]\tikzlastnode.north) edge ([yshift=3mm]\tikzlastnode.south)
        ([xshift=3mm]\tikzlastnode.west) edge ([xshift=-3mm]\tikzlastnode.east)
        }
    }
    }
    \tikzset{KRON/.style={draw,circle,append after command={
        [shorten >=\pgflinewidth, shorten <=\pgflinewidth,]
        (\tikzlastnode.north west) edge (\tikzlastnode.south east)
        (\tikzlastnode.south west) edge (\tikzlastnode.north east)
        }
    }
    }
    \tikzset{SUB/.style={draw,circle,append after command={
        ([xshift=3mm]\tikzlastnode.west) edge ([xshift=-3mm]\tikzlastnode.east)
        }
    }
    }
    \tikzset{VIRTUAL/.style={append after command={
        [shorten >=\pgflinewidth, shorten <=\pgflinewidth,]
        ([xshift=-1mm]\tikzlastnode.west) edge ([xshift=1mm]\tikzlastnode.east)
        }
    }
    }
    \tikzset{line/.style={draw, -latex',shorten <=1bp,shorten >=1bp}}
    \resizebox{0.7\columnwidth}{!}{\begin{tikzpicture}[
        >=stealth,
        node distance=1.5cm,
        auto, thick,
        scale=0.3
        ]
        \node [inverse_dynamics, text width=6em] (inverse_dynamics) {\textbf{Inverse Dynamics Model}};
        \node [KRON, below of=inverse_dynamics] (Kp) {};
        \node [, below= 0.2 cm of Kp] (kp_text) {$K_p$};
        \node [PLUS, right of=Kp,scale=1.2] (plus_1) {};
        \node [below = 2.5 cm of Kp, minimum width=9em] (feed_back_label) {\textbf{\hspace{5mm}Feedback}};
        \node [VIRTUAL, below = 1.0cm of feed_back_label] (feedback_2) {};
        \node [VIRTUAL, below = 0.5cm of feed_back_label] (feedback_1) {};
        \node [PLUS, right of=plus_1, scale=1.2] (plus_2) {};
        \node [below = 2.8 cm of plus_1] (feed_back_corner) {};
        \node [robot, right of=plus_2] (robot) {\textbf{Robot}};
        \node [right of=robot, minimum height=4em] (output) {$\dot q$};
        \node [KRON, below of=Kp] (Kv) {};
        \node [, below= 0.2 cm of Kv] (kv_text) {$K_v$};
        \node [SUB, left = 0.7 cm of Kp] (sub_1) {};
        \node [SUB, left = 0.3 cm of Kv] (sub_2) {};
        \node [,left = 1.0 cm of sub_1] (q_des) {$q_d$};
        \node [,left = 1.4 cm of sub_2] (qd_des) {$\dot q_d$};
        \node [,above = 1.2 cm of q_des] (qdd_des) {$\ddot q_d$};
        \node [,below = 0.2 cm of qd_des] (plan_n) {\textbf{Plan}};
        \node [VIRTUAL, above = 2.5cm of plus_2] (feedback_3) {};
        \node [VIRTUAL, above = 3cm of plus_2] (feedback_4) {};

        \draw [->] (inverse_dynamics) -| node [right] {$u_{ff}$} (plus_2) ;
        \draw [->] (q_des) -> node [left, name=q_des_node] {} (sub_1) ;
        \draw [->] (qd_des) -> node [right, name=qd_des_node] {} (sub_2) ;
        \draw [->] (qdd_des) -> node {} ([yshift=9mm]inverse_dynamics.west) ;

        \draw [->] (sub_1) -> node {} (Kp) ;
        \draw [->] (Kp) -> node {} (plus_1) ;
        \draw [->] (plus_1) -> node {$u_{fb}$} (plus_2) ;
        \draw [->] (plus_2) -> node {$\tau$} (robot) ;
        \draw [->] ([yshift=-20mm]robot.north east) -> node [above, name=q] {$q$} ([yshift=-20mm]output.north west) ;
        \draw [->] ([yshift=20mm]robot.south east) -> node [above, name=qd] {} ([yshift=20mm]output.south west) ;
        \draw [->] ($ (q.south) + (0mm,0mm) $) |- (feedback_1) -| (sub_1.south);
        \draw [->] ($ (output.south) + (0mm,14mm) $) |- (feedback_2) -| (sub_2.south);
        \draw [->] (q) |- (feedback_3) -| ([xshift=4mm]inverse_dynamics.north) ;
        \draw [->] ($(output) + (0mm,10mm) $) |- (feedback_4) -| ([xshift=-4mm]inverse_dynamics.north) ;
        \draw [->] (sub_2) -> node {} (Kv) ;
        \draw [->] (Kv) -| node {} (plus_1) ;
        \begin{pgfonlayer}{background}
            \node [feed_back] [fit = (sub_1) (feed_back_corner)] (feed_back) {};
            \node [plan] [fit = (qdd_des) (plan_n)] (plan) {};
        \end{pgfonlayer}
            
    \end{tikzpicture}}
\caption{\small In this work we learn the model used within an inverse dynamics controller. The controller accepts a target position $q_{target}$ and outputs a plan - the desired joints, joint velocities and joint accelerations  $Q_d = \{q_d, \dot q_d, \ddot q_d\}$ for each time step $t$ of the trajectory. At each time step the inverse dynamics model accepts the current joint position $q$ and velocity $\dot q$, as well as the desired acceleration $\ddot q_d$ and outputs a feed forward torque. The feedback controller computes a feedback torque from the current joints and joint velocities  $q$ and $\dot q$. The combination of $u_{ff}$ and $u_{fb}$ gives us the torque control parameters $\tau$ required to drive the robot arm to $q_{target}$. } 
\label{fig:Controller}
\vspace{-0.4cm}
\end{figure}

A typical model-based control loop for inverse dynamics control as used in our work is shown in Figure \ref{fig:Controller}. The controller accepts a target position $q_{target}$ and outputs a plan - the desired joints, joint velocities and joint accelerations  $Q_d = \{q_d, \dot q_d, \ddot q_d\}$ for each time step $t$ of the trajectory. At each time step, the inverse dynamics model takes the current joint position $q$ and velocity $\dot q$, as well as the desired acceleration $\ddot q_d$ as input and outputs a feed forward torque $u_{ff}$. The inverse dynamics controller then combines $u_{ff}$ with a feedback $u_{fb}$ component to produce the total desired torque $\tau$ applied on the robot. The feedback component is computed through a PD-control law that corrects for the errors caused due to an inaccurate dynamics model: $u_{fb} = K_p(q_d - q) + K_d(\dot{q_d} - \dot{q})$. $u_{fb}$ corrects for inaccuracies in the dynamics model, and pushes the robot towards the desired trajectory. Ideally, for compliant and fast motion, we want small contributions from the feedback, and most of the torque to come from $u_{ff}$, motivating the need for learning accurate inverse dynamics models.
\subsection{Learning Inverse dynamics}
In this work we apply meta-learning to the problem of improving inverse dynamics models, in an attempt to train as quickly as possible. The problem of inverse dynamics learning is widely studied in robotics literature, ranging from learning analytical rigid-body models, as in \cite{atkeson_1986_inertial_param_est} to over-parametric neural-network models, as in \cite{hitzlerlearning}. 
Given the computational and data constraints of learning dynamics models on real robots, there has been significant work on computationally efficient methods for learning and adapting inverse dynamics models \cite{hitzlerlearning,vijayakumar1997local,schaal2002scalable,nguyen2009local,gijsberts2013real}, including learning \textit{incrementally} in an online manner \cite{sigaud2011line,camoriano2016incremental,meier2014incremental,jamone2014incremental}.
Another line of work assumes a prior model (e.g. analytical), which will be inaccurate in the real world, and attempts to learn a model over this error instead of the entire inverse dynamics model  \cite{ratliff2016doomed, meier2016towards, meier2016drifting, nguyen2010using}. In all of these approaches, the dynamics are learned to optimize the mean-squared error, and do not benefit from a state-dependent loss function that can lead to faster learning online. We present a meta-learning framework than can learn structured, state-dependent loss functions for inverse dynamics learning, significantly improving online learning speed.

\subsection{Meta-Learning}
Learning-to-adapt can be considered a form of meta-learning, where an agent attempts to learn \emph{how to learn} from observations. Meta-learning has become very popular within the machine-learning community \cite{finn2017model, meier2018online, bechtle2019meta} and we can differentiate roughly between three types of approaches for learning-to-adapt: Methods that meta-learn initialization of models from which an agent can quickly adapt to new scenarios \cite{finn2017model},  methods that learn an optimizer representation \cite{andrychowicz2016learning, li2016learning}, and finally methods that learn loss functions \cite{bechtle2019meta, epg,vilalta2002perspective}. This work focuses on the latter, learning loss functions.  For a more detailed overview of meta-learning approaches see \cite{hospedales2020meta}.

In this work, we focus on learning loss functions, and build on the work by \cite{bechtle2019meta}, which meta-learns a loss function in order to generalize across tasks. We modify their framework for training loss functions for inverse dynamics learning, and use it to efficiently update dynamics models at test time. 
We additionally introduce state-dependent losses, that allow for changes in the loss landscape based on the current robot state.

\subsection{Meta-Learning Dynamics Models}
 There has also been research in meta-learning with a focus on meta-learning dynamics models, either with fast fine-tuning \cite{nagabandi2018deep} or with online learning of a memory learning rates \cite{meier2018online}. \cite{nagabandi2018deep} meta-learn a dynamics model initialization in simulation that can quickly be fine-tuned during test time to adapt to different settings. However, adaptation to completely novel dynamics, not captured in the meta-train task-distributions, remains a challenge for such approaches. On the other hand, \cite{meier2018online} present an approach to learn the learning rate used in model updates online, which can increase online adaptation efficiency. But because the model that predicts learning rates is learned online, it is affected by forgetting, thus information about prior tasks is lost. 
 
 In this work, we learn a \emph{state-dependent} loss function during an offline meta-training phase; and show that it transfers to previously unseen tasks, with novel dynamics (such as additional load at the end-effector) at meta-test time.

\section{Learning loss functions for Inverse Dynamics Models}
In this section we present our approach to learning structured loss functions for inverse dynamics learning. We built on prior work on meta-learning \cite{bechtle2019meta} which has shown that learned loss functions can lead to faster learning for various supervised learning problems. Here, we propose to learn state-dependent loss functions with structure. 

Our loss-learning framework comprises of two phases:\linebreak
\emph{Phase 1 (meta train):} Meta-Training is the process by which we learn our loss function. More specifically, we use data collected from sine motions on the joints to optimize the parameters $\phi$ of our loss function $\learnedLoss$. 
\emph{Phase 2 (meta test):} Meta-testing (which is similar to a typical training phase) is the process by which we train our inverse dynamics model. We use the $\learnedLoss$ with optimal parameters $\phi*$ to train $f_\theta$ on new tasks.
We start out by recapping how to learn inverse dynamics models given any loss function.
\subsection{Learning Models for Inverse Dynamics Control}
We aim to learn a model of the inverse dynamics - i.e given the current joint positions $q$ and velocities $\dot q$, and the next desired joint acceleration $\ddot q_d$, we want to learn a function $f$, parameterized by $\theta$ such that:
\begin{equation*}
    u_{ff} = f_\theta(q, \dot q, \ddot q_d)
\end{equation*}
where $u_\text{ff}$ is the feed-forward torque that should be applied to achieve the desired acceleration $\ddot q_d$ as per the inverse dynamics model. In this work, we model $f$ via a neural network parameterized by $\theta$, and we aim to learn it from a stream of data as quickly as possible. 
This model is trained on a data set  $\mathcal{D}=\{Q_t=(q_t, \dot q_t, \ddot q_{t+1}), \tau_t\}_{t=1}^T$ by minimizing a loss $\taskLoss$ between the predicted control outputs ($u_\text{ff}$) and ground truth torque values ($\tau$).
The inverse dynamics model can be learned via stochastic gradient descent (SGD) by iteratively updating the parameters $\theta$ as follows:
\begin{equation*}
    \theta_\text{new} = \theta - \alpha \nabla_\theta \taskLoss
\end{equation*}
where $\alpha$ is the chosen learning rate. The MSE loss, just like any other typical loss, is a domain-independent loss that can be used for any function approximation problem. We believe that learning loss functions for inverse dynamics learning can lead to losses that update these models more effectively.
\subsection{Loss Function Representations}\label{ss:loss_fun_rep}
Instead of using unstructured neural-network architectures as in \cite{bechtle2019meta} to represent the loss, we propose to learn \emph{structured} loss functions. 
Next, we describe in detail the different loss architectures explored in our work.
\subsubsection{Multi-Layer Perceptron (mlp)}
In \cite{bechtle2019meta} the learned loss is represented as a neural network. Similar to the MSE loss, the learned loss $l = \learnedLoss(\hat{\tau}, \tau)$ takes as input the predicted output $\hat{\tau} = f_{\theta}$ and its corresponding ground truth value $\tau$, and outputs a loss value that computes an effective distance between $f_{\theta}$ and $\tau$. 
The loss architecture is a fully-connected MLP network with 3 layers of 40 neurons each, with a ReLu activation in the hidden layer. A Softplus activation is applied to the output. This loss representation is our baseline.
\subsubsection{Structured loss (structured)}
The \emph{mlp} architecture doesn't take into account the structure of the inverse dynamics learning problem.
 
Here, we propose to use a parameterized loss of the form:
\begin{equation}
    \learnedLoss = \sum_{j=1}^J \phi_j (f_{\theta j} - \tau_j)^2,
    \label{eq:structuredloss}
\end{equation}
where $J$ denotes the number of controllable joints and $\phi_j \in \mathrm{R}^1$ are learn-able parameters, one per joint $j$. This structured learned loss takes into account to the shape of the inverse-dynamics model's outputs (i.e a predicted torque $f_\theta$ per joint), and learns to assign different weights to the different joints. Intuitively, dynamics errors in some joints can cause poor tracking of desired joint acceleration and this loss learns to weigh torque error on crucial joints more heavily than others. Updating more important joints faster likely leads to improvement in online learning speed. 
\subsubsection{State Dependent loss (state-dependent)}
Our proposed structured loss assumes constant learned weights for the whole robot state space. Here, we introduce a state-dependent loss with weights $\phi_j$ that are a function of the current joint state $\{q, \dot q\}$: 
\begin{equation}
    \learnedLoss = \sum_{j=1}^J \phi_j(q, \dot q) (f_{\theta_ j} - \tau_j)^2.
    \label{eq:stateloss}
\end{equation}
where the functions $\phi_j$ are represented learnable as neural networks with hidden layer of size $32$ with ReLUs as activation functions. Our intuition is that a state-dependent loss can improve upon structured loss by adapting to (ie increasing weight near) areas of the state space where the robot experiences static friction.
\floatstyle{spaceruled}
\restylefloat{algorithm}
\begin{algorithm}[t]
\begin{algorithmic}[1]
\footnotesize{
\STATE{Randomly initialize $\phi$}

\FOR{each $epoch$}
\FOR{each $batch$ in $epoch$}
\STATE{Randomly initialize $\theta$ }
\STATE{Sample a batch of continuous data $q_{t:t+K}, \dot{q}_{t:t+K}, \ddot{q}_{t:t+K}, \tau_{t:t+K}$ }
\STATE{Split batch into 2 batches}
\STATE{$q_A, \dot{q}_A, \ddot{q}_A, \tau_A = q_{t:t+\frac{K}{2}}, \dot{q}_{t:t+\frac{K}{2}}, \ddot{q}_{t:t+\frac{K}{2}}, \tau_{t:t+\frac{K}{2}}$}
\STATE{$q_B, \dot{q}_B, \ddot{q}_B, \tau_B = q_{\frac{K}{2}:K}, \dot{q}_{\frac{K}{2}:K}, \ddot{q}_{\frac{K}{2}:K}, \tau_{\frac{K}{2}:K}$}
\FOR{each $i < \text{iters}_\text{max}$}

\STATE{\emph{Calculate $\theta_\text{new}$ from $\learnedLoss$ with the old $\phi$}}
\STATE{$\hskip1.2em{\theta}_{new} \gets {\theta} - \alpha.\nabla_{{\theta}}\learnedLoss(f_{{\theta}}(q_A, \dot{q}_A, \ddot{q}_A), \tau_A)$}

\STATE{\emph{Update $\phi$ based on $f_{\theta_\text{new}}$'s performance}}
    \STATE{\hskip1.2em $\phi \gets \phi - \eta.\nabla_\phi \taskLoss(f_{{\theta}_\text{new}}(q_B, \dot{q}_B, \ddot{q}_B), \tau_B)$}

\ENDFOR
\ENDFOR
\ENDFOR
}
\end{algorithmic}
\caption{\strut Learning Loss functions at (\textit{meta-train})}
\label{algo:loss-learning-meta-train}
\end{algorithm}
\begin{figure*}[t!]
    \centering
    \begin{subfigure}[b]{0.23\textwidth}
    \includegraphics[width=1.0\textwidth]{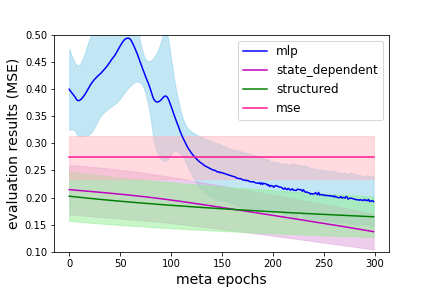}
    \subcaption{meta-train}
    \end{subfigure}
    \begin{subfigure}[b]{0.23\textwidth}
    \includegraphics[width=1.0\textwidth]{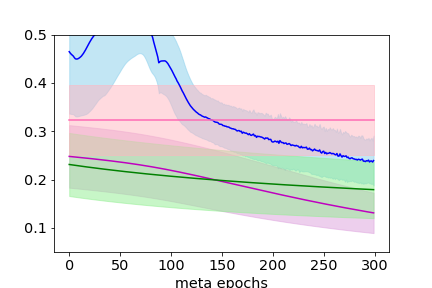}
    \subcaption{meta-test}
    \end{subfigure}
    \begin{subfigure}[b]{0.23\textwidth}
    \includegraphics[width=1.0\textwidth]{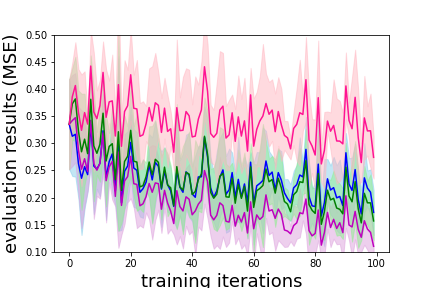}
    \subcaption{train}
    \end{subfigure}
    \begin{subfigure}[b]{0.23\textwidth}
    \includegraphics[width=1.0\textwidth]{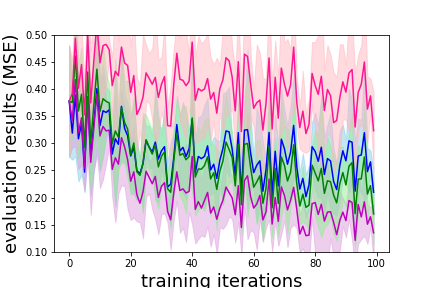}
    \subcaption{test}
    \end{subfigure}
    \caption{\small \textbf{Simulation meta-training phase}: (a) and (b) plot the performance of the learned losses (on meta train data(a) and meta test data(b) collected on hardware) as a function of meta-training epochs. The losses improve over a course of 300 epochs. (c) and (d) show training curves for learning $f_\theta$ when using the final meta (learned) losses and the fixed MSE loss. The state-dependent loss is visibly more stable to train and faster to optimize the task model with compared to other learned losses. Results averaged across 5 seeds.}
    \label{fig:simulation_meta_training_results}
\end{figure*}
\subsection{Learning loss functions}\label{sec:loss-learning}
These loss functions can be learned offline, during a meta-training phase. We employ stochastic gradient descent (SGD) to update the parameters $\phi$ of the learned loss. Note that at each step of the optimization process, $\phi$ must be updated in a way such that training the model $f_\theta$ with the corresponding $\learnedLoss$ brings us a step closer to obtaining an accurate model of the robot's inverse-dynamics. This can be checked by evaluating the resultant inverse-dynamics model with the $\taskLoss$. 

More concretely, we first update the parameters $\theta$ via the update rule $\theta_{new} \longleftarrow \theta - \alpha \nabla_\theta \learnedLoss$ using batches of continuous (sequential) inverse dynamics data $q_{t:t+K}, \dot{q}_{t:t+K}, \ddot{q}_{t:t+K}, \tau_{t:t+K}$, where $K$ is the batch size, from the meta-train data set. This batch is halved, and the first half is used to update the parameters of the inverse dynamics model using the current state of the learned loss $\learnedLoss$. We denote the new updated parameters as $\theta_\text{new}$.   We can now evaluate this updated model $f_{\theta_\text{new}}$ on the 2nd half of the batch. This mimics the incremental learning at test time, where the $\learnedLoss$ updates $f_{\theta}$ on sequentially incoming data. 

The update of the loss parameters $\phi$, is done implicitly through the evaluation of $f_{\theta_\text{new}}$:
\begin{equation}
    \phi \longleftarrow \phi - \eta.\nabla_\phi\taskLoss(f_{\theta_{new}}(Q), \tau)
    \label{eq:phi_update}
\end{equation}
While it is possible to analytically derive this gradient (See \cite{bechtle2019meta}), one can use modern libraries for such meta-gradient computations. In particular, we use \emph{higher} \cite{grefenstette2019generalized}.
 Our approach for meta-training a learned loss for learning inverse-dynamics models is summarized in Algorithm \ref{algo:loss-learning-meta-train}

\section{Simulation Experiments}
In this section we evaluate our proposed loss function representations for inverse dynamics learning in simulation. In particular, we aim to evaluate how well the different loss function representations train and how online adaptation of inverse dynamics with learned loss compares to using a standard MSE loss.
\subsection{Meta-training: Learning loss functions}\label{ss:sim_meta_train}
We start by training our loss function representations in simulation.
In our meta-training phase, we collect data, meta-learn the loss functions, and evaluate the different types of learned losses (\emph{mlp}, \emph{structured}, and \emph{state-dependent}).
%
\subsubsection{Data Collection}\label{ss:data_collect_sim_meta}
For our simulation experiments, we record the motion of the KUKA iiwa moving in the PyBullet simulator \cite{coumans2016pybullet}. We collect multiple sine-motion trajectories that were generated around the rest posture of the arm. More specifically, let $q_\text{rest}$ be the rest posture, then we generate the desired joint position at time step $t$ to be $q_t = \text{A} \sin{(2 \pi f \Delta_t t)} + q_\text{rest}$, where $\Delta_t$ is the length of a time step, $f$ is the frequency of the sine motion, and $A$ is the amplitude of the sine motion, with a value that is chosen to avoid self-collision.  
To get a rich data set we collect sine-motion data at various frequencies. Specifically, we collect sine motion data at frequencies $f = [0.01, 0.02, \dots, 0.09]$. The simulator runs at 240 Hz, and for each frequency we collect 10s of data. We split the collected data into meta-train and meta-test data sets, where meta-train data consists of data collected with frequencies $[0.01, 0.03, 0.05, 0.06, 0.07, 0.08]$ and meta-test data of data collected with frequencies $[0.02, 0.04, 0.09]$.
\begin{figure*}[t!]
    \centering
    \begin{subfigure}[b]{0.23\textwidth}
    \includegraphics[width=1.0\textwidth]{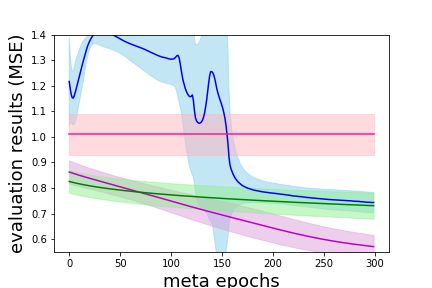}
    \subcaption{meta-train}
    \end{subfigure}
    \begin{subfigure}[b]{0.23\textwidth}
    \includegraphics[width=1.0\textwidth]{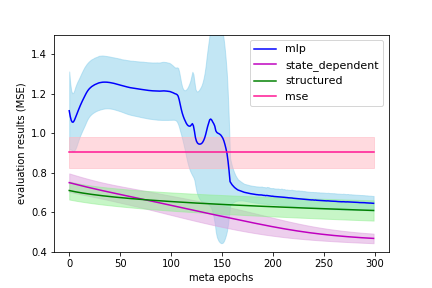}
    \subcaption{meta-test}
    \end{subfigure}
    \begin{subfigure}[b]{0.23\textwidth}
    \includegraphics[width=1.0\textwidth]{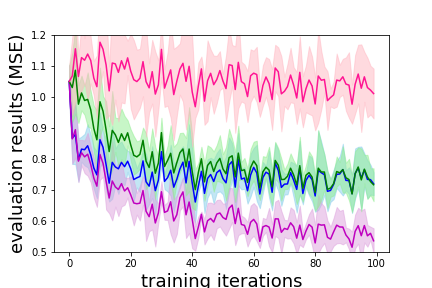}
    \subcaption{train}
    \end{subfigure}
    \begin{subfigure}[b]{0.23\textwidth}
    \includegraphics[width=1.0\textwidth]{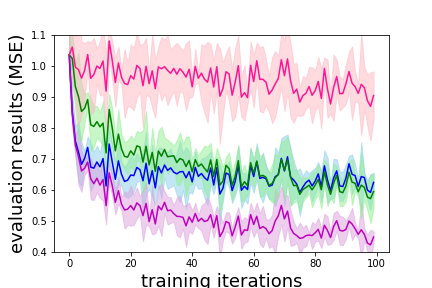}
    \subcaption{test}
    \end{subfigure}

    \caption{\small \textbf{Hardware meta-training phase:}(a) and (b) plot the performance of the learned losses (on meta train data(a) and meta test data(b) collected on hardware) as a function of meta-training epochs. The losses improve over a course of 300 epochs, with 100 gradient steps per epoch. (c) and (d) show training curves for learning $f_\theta$ when using the final meta (learned) and fixed (mse) losses. The state-dependent loss is visibly more stable to train and faster to optimize the inverse-dynamics model with, compared to other learned losses. Results are averaged across 5 seeds.
    } 
    \label{fig:hardware_meta_training_results}
\end{figure*}
\subsubsection{Evaluation of Meta-Training}\label{sec:sim-meta-training-eval}
We first meta-train our learned loss functions \emph{mlp}, \emph{structured} and \emph{state-dependent} described in Section \ref{ss:loss_fun_rep} with the meta-train data set according to Algorithm \ref{algo:loss-learning-meta-train} with hyper-parameters:
\begin{center}
\begin{footnotesize}
    \begin{tabular}{c c c c}
			\toprule
			\bf Batch Size & \bf $\alpha$ & \bf $\eta$ & \bf $\text{iters}_\text{max}$ \\ 
			\midrule
			256 & 0.001 & 0.01 & 10\\
			\bottomrule
	\end{tabular}
\end{footnotesize}
\end{center}
After meta-training has finished, we evaluate our learned losses against Pytorch's built in Mean Squared Error Loss \cite{paszke2019pytorch} for comparison by training an inverse-dynamics model with each of the losses. We perform two kinds of evaluations:

\emph{During meta-training} After each epoch, we use the current losses and use them to train inverse dynamics models, on 1) the meta-train data itself, and on 2) on the held-out meta-test data (see Figure~\ref{fig:simulation_meta_training_results} (a) and (b) respectively. We compare to using the MSE loss. Note while we use the various losses to perform the learning we evaluate the MSE achieved for all of them, to make results comparable. 

\emph{After Meta-Training:} We use all learned losses and the MSE loss to perform a final training of the inverse dynamics on both meta-train and meta-test data, for $100$ gradient steps. Results of these training curves, again visualizing MSE achieved by the various losses, are shown in Figure~\ref{fig:simulation_meta_training_results} (c) and (d), respectively.
    
\subsubsection{Analysis}\label{ss:sim_meta_train_analysis}
We note from Figure \ref{fig:simulation_meta_training_results} (a) and (b), that by 150 epochs, all three of the learned losses have achieved better results than MSE. While the \emph{mlp} loss improves upon MSE loss, it is also less stable to train than the \textit{structured} and \textit{state-dependent} losses (observe how the evaluation metric for the model $f_\theta$ varies smoothly with the number of meta-epochs for these particular losses). This shows that incorporating information about the inverse dynamics model and the nature of its outputs in the loss function make the loss function easier to train. The best performer amongst all the learned losses is the \emph{state-dependent} loss, suggesting that the loss landscape for training inverse-dynamics model is dependent on the current state.

Plots from Figure \ref{fig:simulation_meta_training_results} (c) and (d) show that training the inverse dynamics model via the learned losses progresses much faster as compared to training via MSE loss. Therefore, these results suggest that learned losses indeed lead to faster training of inverse dynamics models.
A key takeaway is that the \textit{structured} loss performs better than the \textit{mlp} loss, and \textit{state-dependent} is the best performer.
\subsection{Online Adaptation}
\begin{figure}[h]
    \centering
  \includegraphics[width=.46\columnwidth]{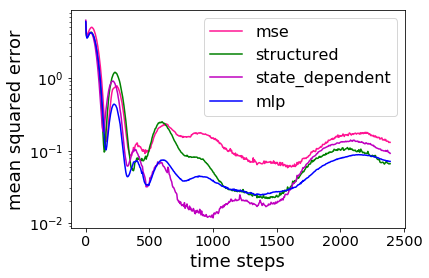}
\label{fig:linersammenhang}
\caption{\small Online Learning with learned losses: MSE of predicted torque of the inverse dynamics model as a function of time. At each time step the model is updated through one gradient step.}
\label{fig:simulator_online_learning}
\end{figure}
In the previous experiments, we evaluated the trained loss functions on learning inverse dynamics models from randomly sampled batches. Here, we evaluate how fast training progresses when using the learned losses to train on streaming data. 
For this experiment, we use the loss functions trained in the previous experiment to train a randomly initialized inverse dynamics model online. We use the meta-test trajectories obtained from the data collection process described in section \ref{ss:data_collect_sim_meta}, but instead of randomly sampling from that data, training passes through the data sequentially, in batches of size 5.  In detail, at every update step $k$, we create a batch from the next 5 time steps, record the prediction error (MSE) made on that batch, and then perform one gradient update step on model parameters $\theta$ based on the learned losses.
We visualize this prediction error as a function of time in Figure \ref{fig:simulator_online_learning}. The slope and stability of the MSE curves in the figure indicate how quickly the model is able to adapt to an incoming stream of data. We note from Figure \ref{fig:simulator_online_learning} that the MSE curves for the inverse model trained via learned losses output lower values throughout the training process than the ones trained using an MSE loss. This indicates that training with a learned loss is much more suitable when the model needs to adapt in an online setting.
\begin{figure*}
	\begin{tabular}{lccccc}
		\toprule
		& Reach &  Lift &  Move Over &  Lower &  Retract\\ 
		\midrule
		\raisebox{-.5\normalbaselineskip}[0pt][0pt]{\rotatebox[origin=c]{90}{\hspace{2.5cm}}} &
		{\begin{subfigure}[b]{0.13\textwidth}
    \includegraphics[trim=0 126 0 90, clip, width=1.0\textwidth]{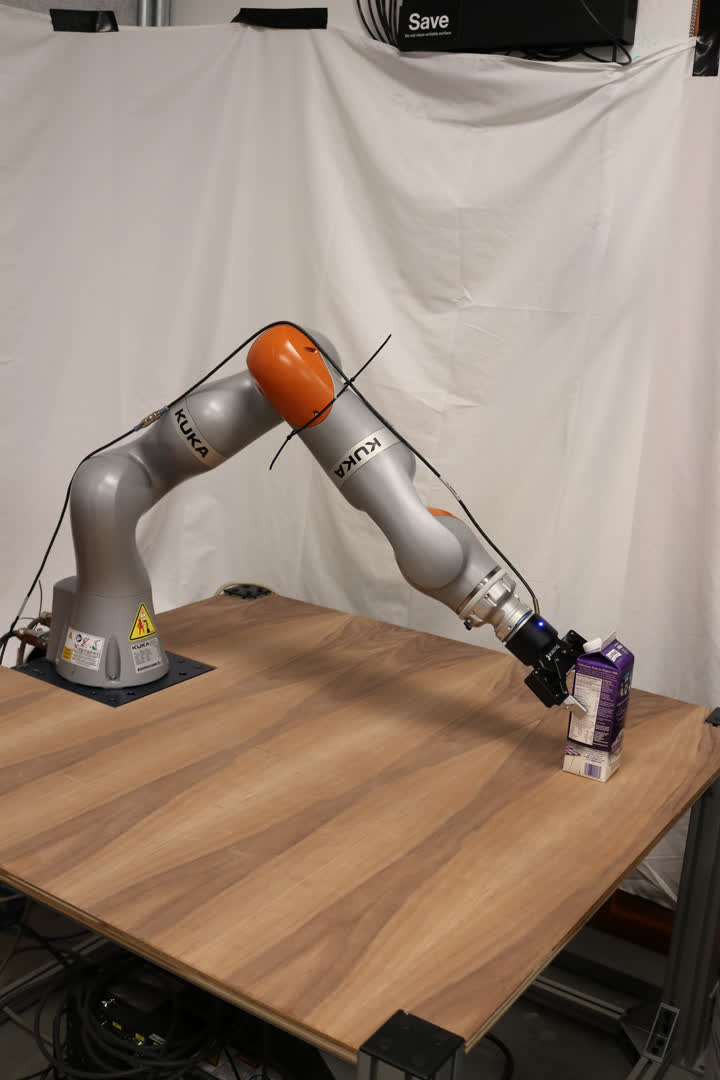}
    \end{subfigure}} &
		{\begin{subfigure}[b]{0.13\textwidth}
    \includegraphics[trim=0 126 0 90, clip, width=1.0\textwidth]{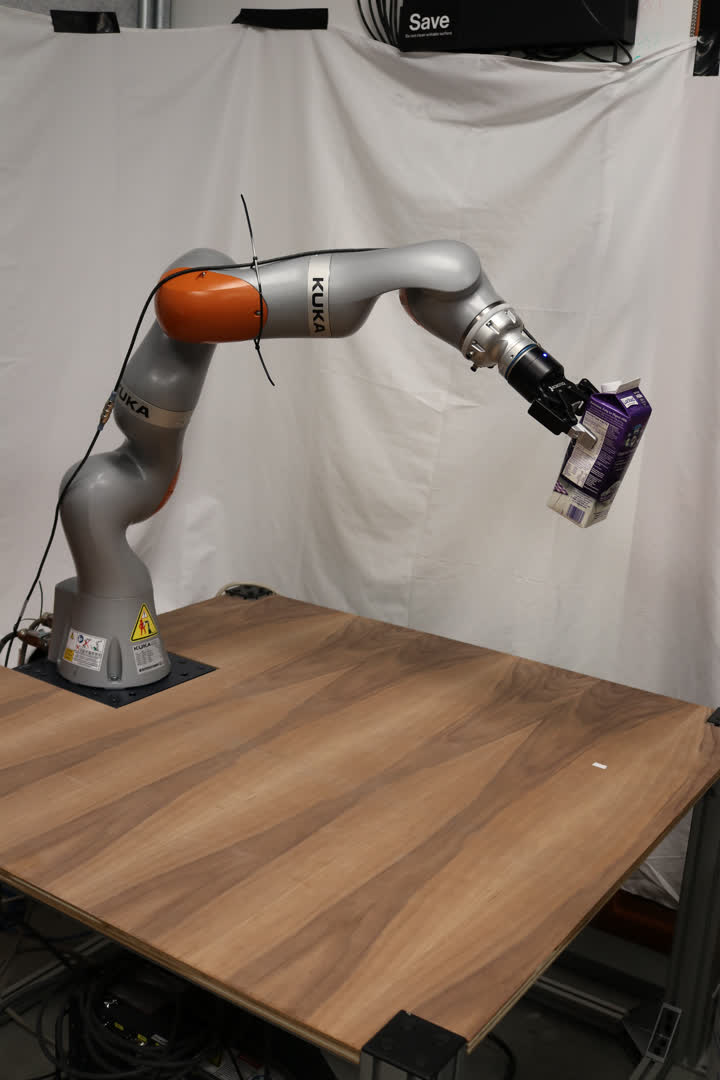}
    \end{subfigure}} &
		{\begin{subfigure}[b]{0.13\textwidth}
    \includegraphics[trim=0 126 0 90, clip, width=1.0\textwidth]{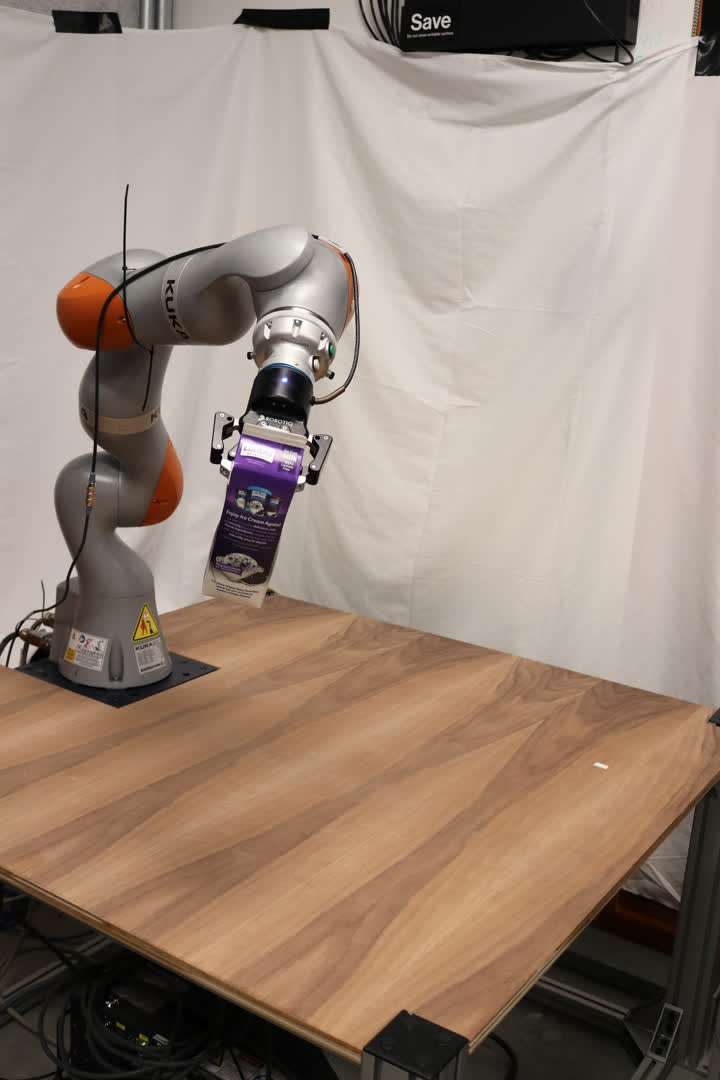}
    \end{subfigure}} &
		{\begin{subfigure}[b]{0.13\textwidth}
    \includegraphics[trim=0 126 0 90, clip, width=1.0\textwidth]{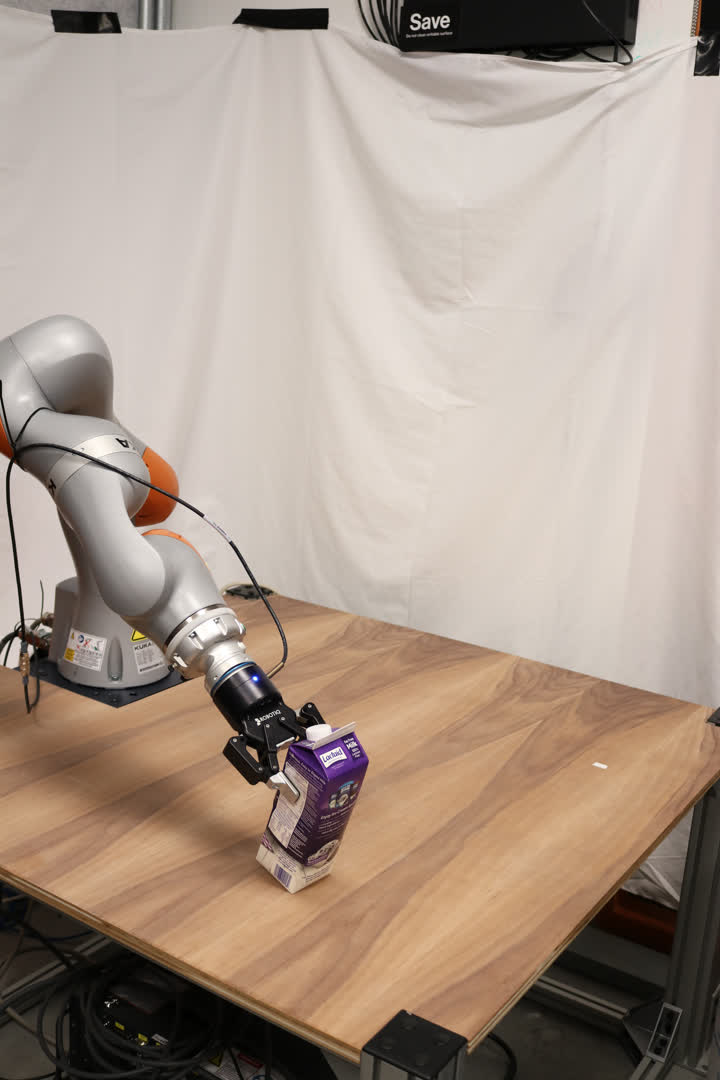}
    \end{subfigure}} &
    {\begin{subfigure}[b]{0.13\textwidth}
    \includegraphics[trim=0 126 0 90, clip, width=1.0\textwidth]{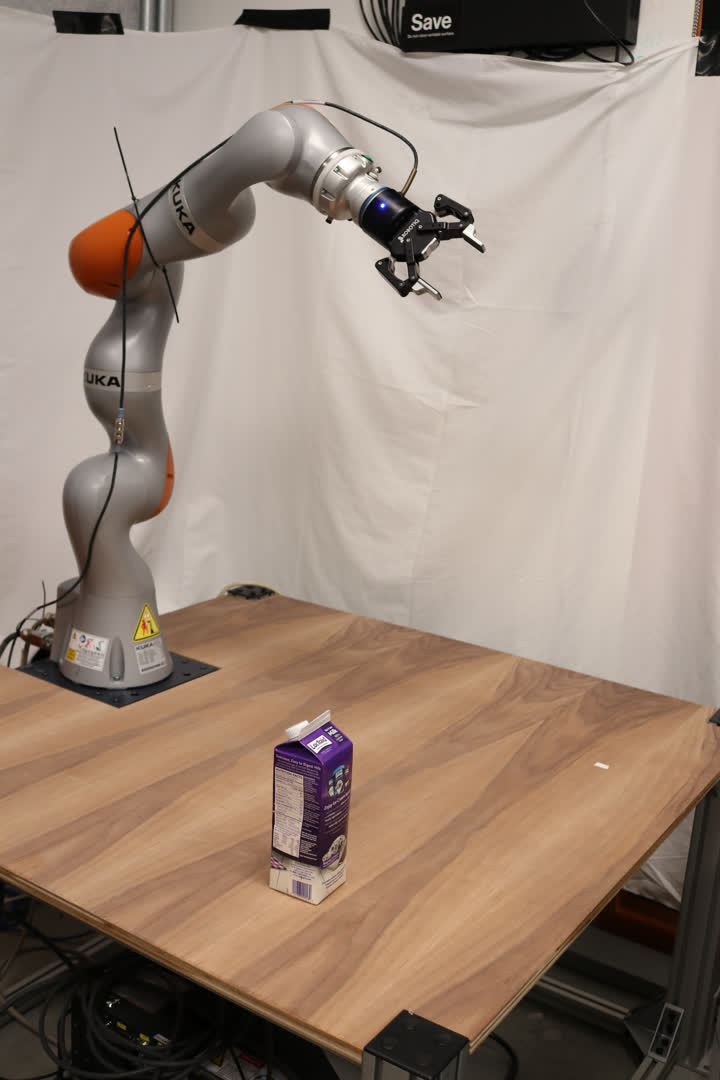}
    \end{subfigure}}\\
		\raisebox{-.5\normalbaselineskip}[0pt][0pt]{\rotatebox[origin=c]{90}{\hspace{2.5cm}{Sim2Real}}} &
		{\begin{subfigure}[b]{0.165\textwidth}
	\includegraphics[width=1.0\textwidth]{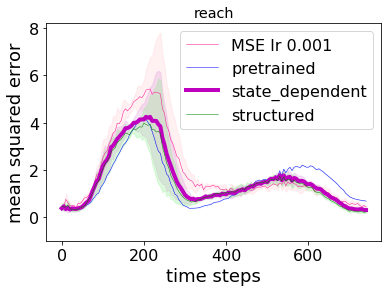}
    \end{subfigure}} &
		{\begin{subfigure}[b]{0.165\textwidth}
    \includegraphics[width=1.0\textwidth]{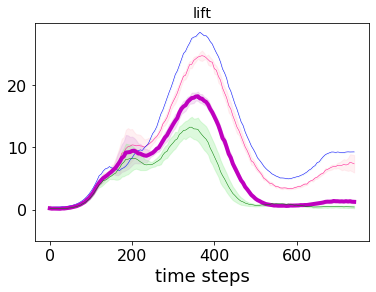}
    \end{subfigure}} &
		{\begin{subfigure}[b]{0.165\textwidth}
    \includegraphics[width=1.0\textwidth]{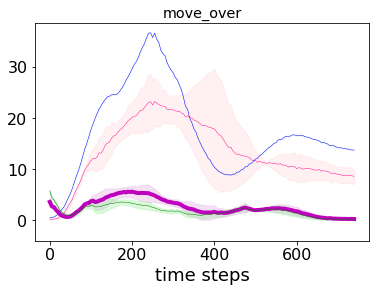}
    \end{subfigure}} &
		{\begin{subfigure}[b]{0.165\textwidth}
    \includegraphics[width=1.0\textwidth]{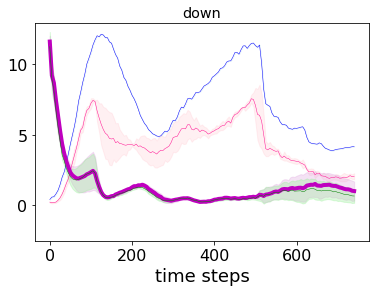}
    \end{subfigure}} &
    {\begin{subfigure}[b]{0.165\textwidth}
    \includegraphics[width=1.0\textwidth]{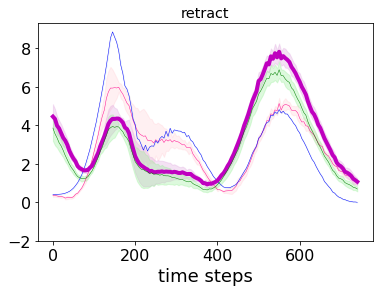}
    \end{subfigure}}\\
		\begin{turn}{90}
			 \hspace{0.4cm}Real2Real
		\end{turn} &
		{\begin{subfigure}[b]{0.165\textwidth}
	\includegraphics[width=1.0\textwidth]{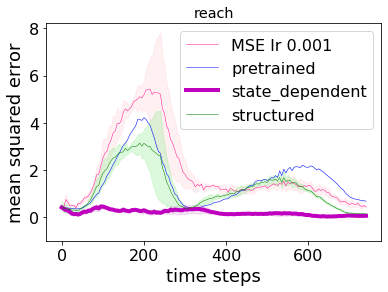}
    \end{subfigure}} &
		{\begin{subfigure}[b]{0.165\textwidth}
    \includegraphics[width=1.0\textwidth]{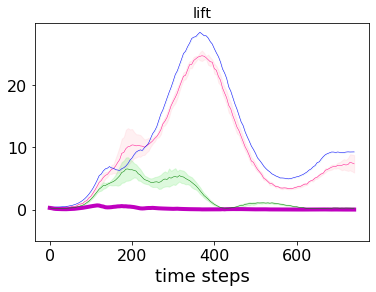}
    \end{subfigure}} &
		{\begin{subfigure}[b]{0.165\textwidth}
    \includegraphics[width=1.0\textwidth]{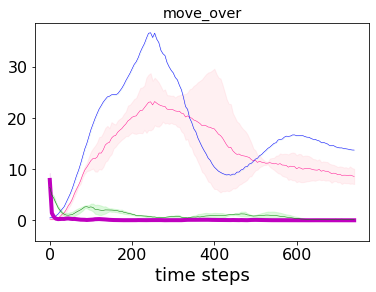}
    \end{subfigure}} &
		{\begin{subfigure}[b]{0.165\textwidth}
    \includegraphics[width=1.0\textwidth]{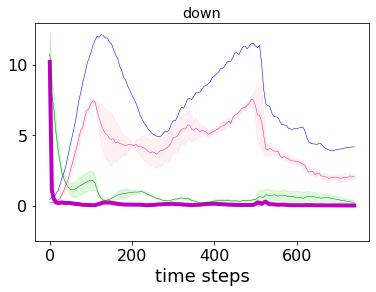}
    \end{subfigure}} &
    {\begin{subfigure}[b]{0.165\textwidth}
    \includegraphics[width=1.0\textwidth]{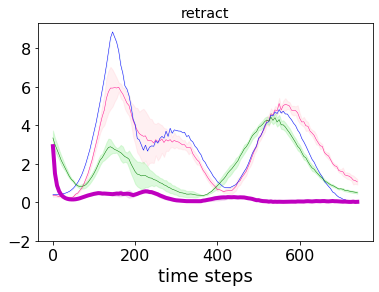}
    \end{subfigure}}\\
        \begin{turn}{90}
			 \hspace{0.1cm}Real2Real 2
		\end{turn} &
		{\begin{subfigure}[b]{0.165\textwidth}
    \includegraphics[width=1.0\textwidth]{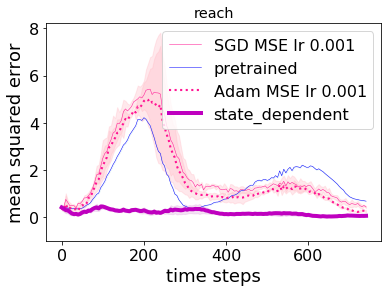}
    \end{subfigure}} &
		{\begin{subfigure}[b]{0.165\textwidth}
    \includegraphics[width=1.0\textwidth]{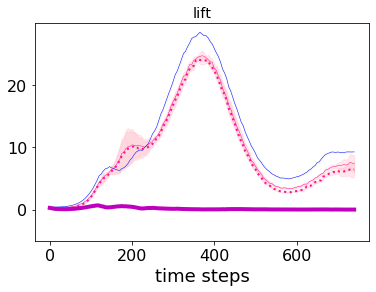}
    \end{subfigure}} &
		{\begin{subfigure}[b]{0.165\textwidth}
    \includegraphics[width=1.0\textwidth]{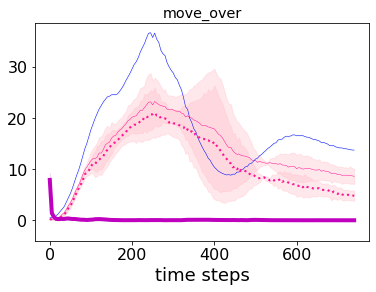}
    \end{subfigure}} &
		{\begin{subfigure}[b]{0.165\textwidth}
    \includegraphics[width=1.0\textwidth]{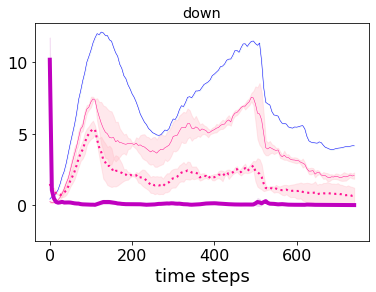}
    \end{subfigure}} &
    {\begin{subfigure}[b]{0.165\textwidth}
    \includegraphics[width=1.0\textwidth]{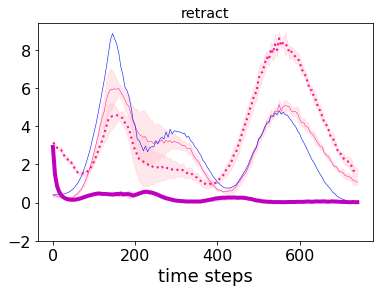}
    \end{subfigure}}\\
    
	\end{tabular}
	\caption{\small The 1st row depicts a snapshot while executing each of the pick and place motions. 2nd to 4th row: Performance (MSE) of $f_\theta$ on the 5 motions as it is being trained using $\learnedLoss$ or $\taskLoss$. 2nd row (Sim2Real) uses loss functions trained in simulation, while the 3rd and 4th row (Real2Real) uses loss functions trained on hardware data.  \emph{Sim2Real, row 2:} The learned losses adapt the models faster/better then the MSE loss; however the \emph{structured} and \emph{state-dependent} loss perform similarly well (on average). \emph{Real2Real, 3rd row:} The \emph{state-dependent} loss significantly outperforms all other losses. \emph{Real2Real, 4th row:} The \emph{state-dependent} loss also significantly outperforms an adaptive optimizer. Also note the blue line: this line represents a dynamics model pretrained on all meta train data with a normal MSE loss and used as is (not adapted). Notice that the pretrained model does not perform well on this test task, highlighting the need for adaptation to this novel data distribution.}
	\label{fig:hardware_online_results}
\end{figure*}

\section{Hardware experiments}
We now turn to evaluating our proposed loss architectures on real hardware data. While simulation results are already promising, we expect the benefits of using \emph{state-dependent} losses to be even larger on real data, as a lot of effects like stiction do not appear in simulation.
\subsection{Meta-training: Learning loss functions}
Similar to simulation in Section~\ref{sec:sim-meta-training-eval}, we first evaluate the meta-training phase - in particular, how well our learned losses can optimize inverse dynamics models on unseen data. 
\subsubsection{Data Collection} \label{ss:data_collect_real_meta}
For our real hardware experiments, we collect two types of data sets from our Kuka iiwa, which is gravity compensated. Similar to our simulation experiments, we collect sine motion data at frequencies $f = [0.02, 0.03, \dots, 0.08]$. Our controller runs at 250 Hz, and for each frequency we collect 30s of data. We split the collected data into meta-train and meta-test data sets, where meta-train data consists of data collected with frequencies $[0.02, 0.03, 0.04, 0.06, 0.08]$ and meta-test data of data collected with frequencies $[0.05, 0.07]$.

For meta-training we use the same hyper-parameters as we did in Section \ref{ss:sim_meta_train}

\subsubsection{Evaluation and Analysis}\label{ss:real_meta_train_analysis}
\vspace{2mm}
The evaluation statistics plotted in Figure \ref{fig:hardware_meta_training_results} and our analysis for evaluating meta-training for the real data from them mirror those for the simulation data (Section \ref{sec:sim-meta-training-eval}). 
Figures \ref{fig:hardware_meta_training_results} (a) and (b) demonstrate that meta-training learned losses with a structure, (i.e, both \emph{structured} and \emph{state-dependent} losses) is much more stable than meta-training the \emph{mlp }learned loss. From Figures \ref{fig:hardware_meta_training_results}(c) and (d), we again note that inverse-dynamics models train much faster when learned-losses are used for optimization as opposed to MSE loss in the order: \textit{state-dependent} $>$ \textit{structured} $>$ \textit{mlp}.

Note that while in the corresponding experiment on the simulation (Section \ref{ss:sim_meta_train}), the \emph{state-dependent} loss is the best performer, its improved performance on the real data is much more visible. 

This significant improvement of the state-dependent loss, when compared with the experiments on simulation, can be explained by considering the fact that the hardware has to deal with real world issues, such as static friction and noise. We believe the state-dependent loss has learned how to penalize errors depending on where in the state-space the robot is. 

\subsection{Online Adaptation on Pick and Place Task}
We now use the learned losses to online learn the inverse dynamics model of a pick and place task.
\begin{table*}[h]
		\centering
		\begin{center}
		
		\vspace{1mm}
			\begin{tabular}{l c c c c c c c c c c}
				\toprule
				\bf Loss Functions & \multicolumn{2}{c}{\bf Reaching} & \multicolumn{2}{c}{\bf Lifting} & \multicolumn{2}{c}{\bf Moving Over} & \multicolumn{2}{c}{\bf Putting Down} & \multicolumn{2}{c}{\bf Retracting}\\ 
				& Mean & Std & Mean & Std & Mean & Std & Mean & Std & Mean & Std \\
				\midrule
				mse, lr 0.001 & 1.881 & 1.588 & 9.253 & 7.191 & 13.535 & 7.399 & 4.242 & 1.934 & 2.705 & 1.685\\
                mse, lr 0.01 & 1.602 & 1.374 & 6.53 & 6.243 & 3.624 & 3.129 & 1.376 & 1.642 & 3.559 & 2.271\\
                structured & 1.24 & 0.958 & 2.144 & 2.251 & 1.083 & 0.985 & 0.885 & 1.345 & 1.811 & 1.199\\
                state dependent & \textbf{0.196} & \textbf{0.142} & \textbf{0.152} & \textbf{0.174} & \textbf{0.136} & \textbf{0.661} & \textbf{0.156} & \textbf{0.849} & \textbf{0.237} & \textbf{0.311}\\
                mlp & 1.203 & 1.254 & 6.124 & 4.844 & 9.168 & 5.324 & 3.647 & 1.649 & 2.328 & 1.812\\
				\bottomrule
			\end{tabular}
		\end{center}
		\caption{\label{tab:real2real} \small Online adaptation performance (MSE) of inverse-dynamics model on pick and place task. For each motion segment, results are averaged across 3 trials, and 5 random seeds per trial. The state-dependent loss is the performer across each motion trajectory both in terms of converging to the lowest error on an average as well as having the least amount of variance in terms of prediction performance.}
\end{table*}

\vspace{1mm}
\subsubsection{Data Collection and Setup}
The second data set consists of data collected for a pick-and-place task (see top row of Figure~\ref{fig:hardware_online_results}). We collect 3 trials of picking-and-placing a milk carton, which weighs $857$ grams, and separate the task into 5 motions: reaching for the carton (\emph{reach}), lifting the carton (\emph{lift}), moving the carton across the table (\emph{move-over}), putting the carton down (\emph{move-down}), and retracting the arm to it's rest posture (\emph{retract}).

For each trial, the inverse-dynamics model $f_\theta$ is initialized randomly and then trained by online updating its parameters $\theta$ using the various learned losses on the stream of sequential data points $\{q_t, \dot q_t, \ddot q_{t+1}, \tau_t\}$ we receive while undergoing each of the above 5 motions. We train our model $f_\theta$ on each of the 5 motion phases sequentially, always warm-starting the model for the next motion. 
As a baseline, we also pretrain a dynamics model $f$ on the meta-train data. In the following evaluation this pretrained model is not adapted online, and thus is an indicator on how much adaptation is needed.
\subsubsection{Evaluation}
We record the MSE between predicted and actual torques while training on each task within pick and place with the learned losses \emph{mlp}, \emph{structured} and \emph{state-dependent} each with learning rates 0.001. For comparison, we plot these in Figure \ref{fig:hardware_online_results} alongside an MSE loss (with learning rate .001) and with our baseline pretrained model. The second row of Figure \ref{fig:hardware_online_results} shows the results for learned losses trained on the simulator data, indicating the ability of a sim-to-real transfer from the losses meta-trained on sim data to the model-training on real data. For the third row of the same Figure, we train the learned losses on data collected from a sine-motion task on real hardware. In the last row of Figure \ref{fig:hardware_online_results}, we compare against an adaptive optimizer (Adam). We see that a SGD optimizer optimizing our state dependent loss outperforms an adaptive optimizer that optimizes an MSE loss, as well. Overall, we make 2 observations: 1) all learned losses outperform the standard MSE loss; 2) we see a clear improvement over the losses trained in simulation; most strikingly the state-dependent loss significantly outperforms all other losses. It is able to achieve low error predictions throughout all motions.  
\vspace{2mm}
\subsubsection{Analysis}
Three separate conclusions emerge from the analysis of the results in Figure \ref{fig:hardware_online_results}:

i) \emph{Adaptation is needed}: The pretrained model (pre-trained on meta-train data) does not perform well on this test task. This shows that the data distribution is significantly different from the meta-train data distribution and adaptation is needed.

ii) \emph{Sim-to-Real Transfer}: 
For most of the executed motions, models trained by optimizing the losses learned on the simulator data seem to perform better than models trained with MSE losses, especially when the same $\alpha$ from  meta-training (Algorithm \ref{algo:loss-learning-meta-train}) is used to train them. This indicates that the learned losses can be transferred from simulations to real data to some extent. 

iii) \emph{State-dependent losses adapt better}:
Models optimized with learned losses perform consistently better than those optimized using the MSE loss at the same learning rate - 0.001. Specifically, the state-dependent loss outperforms all other losses (including the model trained with MSE loss at a higher learning rate). This can be seen in Table \ref{tab:real2real}, as state dependent loss has the lowest error on average and lowest variance in terms of prediction performance for each motion. Consequently, these results suggest that keeping track of the state allows our learned loss to better adapt the dynamics model (for example, in cases of stiction in our robot joints).  Additionally, we observe that the evaluation curves for the state-dependent loss are much smoother and show less variance across multiple trajectories (which follow the same motion as we progress through training $f_\theta$). This indicates that training a model with state-dependent loss on an online stream of data is significantly more stable than training with other learned-losses or MSE. 
\begin{figure}[h]
    \centering
    \begin{subfigure}[b]{0.23\textwidth}
    \includegraphics[width=1.0\textwidth]{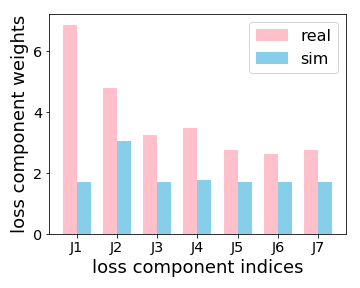}
    \end{subfigure}
    \caption{ \small This plot depicts the $\phi$ values for $structured$ loss trained on robot (real) and simulator (sim) data (loss updated in equation \ref{eq:structuredloss}). The difference between $\phi$s for these data sets is likely because real data contains examples of non-ideal motion patterns with static friction and noise. }

    \label{fig:phi}
\end{figure}
\begin{figure}[h]
    \centering

    \begin{subfigure}[b]{0.43\textwidth}
    \includegraphics[width=1.0\textwidth]{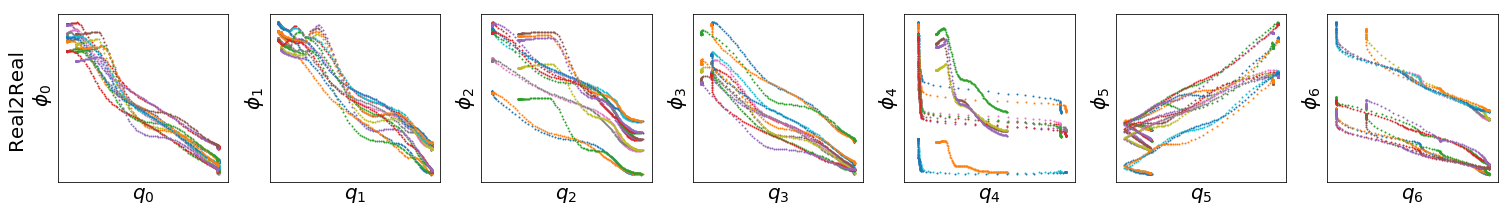}
    \end{subfigure}
    \begin{subfigure}[b]{0.43\textwidth}
    \includegraphics[width=1.0\textwidth]{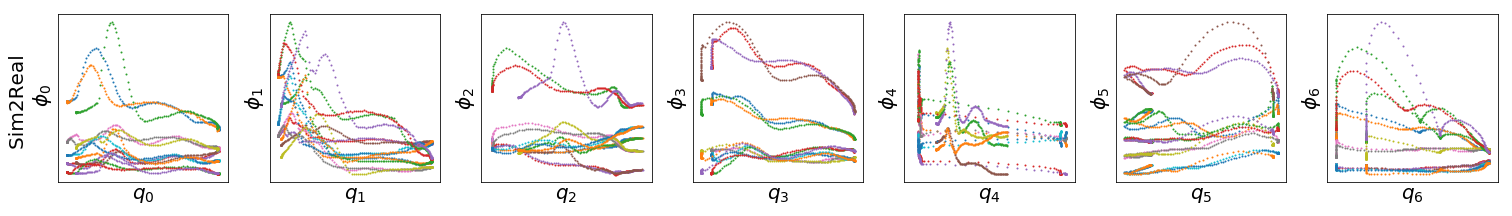}
    \end{subfigure}
    
    \caption{\small We show scatter plots of the $\phi$ against joint angles $q$ for the \emph{state-dependent} loss (equation \ref{eq:stateloss}) trained on robot (1st row) and simulator (2nd row) sine-motion data for multiple trajectories (differently colored) of the \emph{reaching} motion on the robot. The correlation between $\phi$ and $q$ demonstrates how losses change as a function of the state.} 
    \label{fig:phi_state_dep}
\end{figure}

\subsection{Analysis of Structured Losses}
Our structured losses (\emph{structured} and \emph{state-dependent}) incorporate the notion that torques at some of the joints are more informative than others when measuring a distance between torque values and therefore learn to weigh each component of the loss function that corresponds to a torque dimension separately. We visualize the weights of the \emph{structured} loss in Figure~\ref{fig:phi}, both for the loss trained in simulation as well as the loss trained on hardware. Note that the distribution of weights differs between the simulation and hardware data. We postulate that the structured loss learned on real robot data captures some of the reality gap between hardware and simulation in its weight distribution, therefore adapting to what we might only see in a hardware experiment. It is possible that all joints have a higher weight for the real data because there exists some stiction on the joints, causing them to require higher torque values to move the same distance as in simulation. 
We also visualize the weight of the first joint of the state-dependent loss (Figure \ref{fig:phi_state_dep}) as a function of the joint state. We can clearly see that the weights on the joints vary as a function of the state. We believe that it is highly likely that keeping track of state captures more of these real world issues than structured loss does, allowing us to have even better performance.

\section{Conclusion}
In this work we contribute a framework for applying meta-learning to adaptation of inverse dynamics models and improve upon previous meta-learning loss \cite{bechtle2019meta} methods by adding structure and state dependency to the learned loss, resulting in better adaptation to online data. 
Our evaluation shows that all learned losses perform better than our baseline (MSE) on training an inverse dynamics model. However, specifically adding structure and state-dependency to the learned losses produces faster results with better (and more stable) adaptation to hardware data. This result provides experimental evidence for our intuition that a state-dependent loss can capture non-linear and unmodelled effects such as static friction.

We see several promising directions for further research involving the methods we have introduced in this work. Currently, due to safety concerns, the online adaptation experiments on the hardware were performed using data collected offline from the robot, but fed to the model sequentially, mimicking an actual online setting. In future, we plan on investigating methods to safely apply meta-learning directly on robots. 

Another intriguing direction would be to investigate the \emph{state-dependent} loss further in order to verify some of our hypotheses relating to its exceptional performance. For instance, we would like to demonstrate and visualize the inner workings of the \emph{state-dependent} loss. In addition, we hope to explore how structured and state-dependent meta losses improve learning of other types of models for inverse dynamics - such as  analytical models \cite{sutanto2020encoding, atkeson_1986_inertial_param_est}. Finally, to dive deeper into the analysis of our meta-learned losses, we intend to compare our loss-learning approach to other meta-learning methods such as \cite{nagabandi2018learning, kaushik2020fast} to understand when a learned loss leads to better adaptation than a meta-trained model.

\bibliographystyle{IEEEtran}
\bibliography{bib/meta_learning,bib/ml3_paper,bib/inverse_dynamics}

\end{document}